\documentclass[]{fairmeta}
\usepackage{tabularx}
\usepackage{makecell}
\usepackage{adjustbox}
\usepackage{amsmath,amssymb}
\usepackage[noabbrev,nameinlink]{cleveref}
\usepackage{comment}
\usepackage{xspace}

\setcitestyle{square,comma,numbers,sort&compress}

\makeatletter
\DeclareRobustCommand\onedot{\futurelet\@let@token\@onedot}
\def\@onedot{\ifx\@let@token.\else.\null\fi\xspace}

\makeatother

\definecolor{rynn}{RGB}{108,92,186}
\renewcommand{\paragraph}[1]{\vspace{1.25mm}\noindent\textbf{#1}}

\newcommand{\huggingface}{\raisebox{-1.5pt}{\includegraphics[height=1.05em]{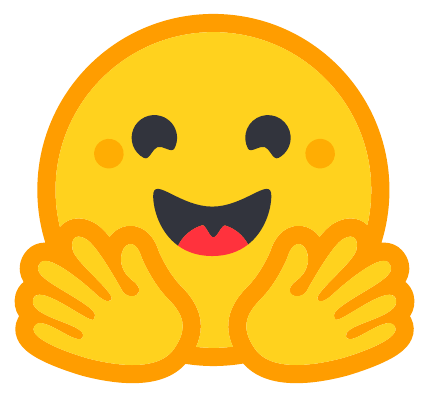}}\xspace}
\newcommand{\github}{\raisebox{-1.5pt}{\includegraphics[height=1.05em]{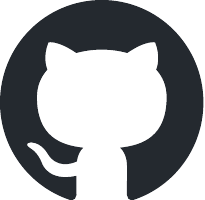}}\xspace}
\newcommand{\homepage}{\raisebox{-1.5pt}{\includegraphics[height=1.05em]{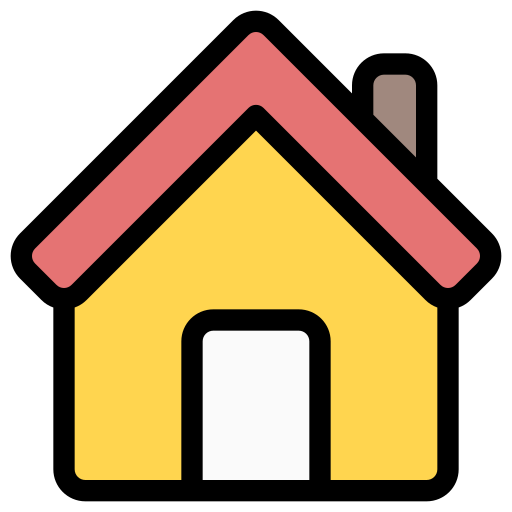}}\xspace}

\title{RynnBrain 1.1: Towards More Capable and Generalizable Embodied Foundation Model\footnote{We would like to thank \href{https://www.wuji.tech/en/}{WUJI TECH} for their generous and unconditional hardware support.}}

\author[*,\dagger]{Kehan Li}
\author[*]{Bohan Hou}
\author[*]{Minghao Zhu}
\author[*]{Tianyi Zhang}
\author[*]{Zesen Cheng}
\author[*]{Zhikai Wang}
\author[*]{Sicong Leng}
\author[*,\dagger]{Xin Li}
\author[*]{Xiao Lin}
\author[*]{Biying Yao}
\author[*]{Minghua Zeng}
\author[*]{Jiangpin Liu}
\author[*]{Ronghao Dang}
\author[*]{Jiayan Guo}
\author{Siteng Huang}
\author{Haoyu Zhao}
\author{Heng Ping}
\author{Yaxi Zhao}
\author{Tong Zhao}
\author{Kexiang Wang}
\author{Tong Lu}
\author{Shengke Xue}
\author{Jiahao Tang}
\author{Yulei Wang}
\author{Zejing Wang}
\author{Jianwei Gao}
\author{Shijian Lu}
\author{Chengju Liu}
\author{Jianfei Yang}
\author{Mingxiu Chen}
\author{Deli Zhao}

\affiliation[1]{DAMO Academy, Alibaba Group}
\affiliation[2]{Hupan Lab}
\contribution[*]{Core contributors}
\contribution[\dagger]{Project lead}

\abstract{
\vspace{-1em}
We present RynnBrain 1.1, a family of embodied foundation models spanning 2B, 9B, and 122B-A10B scales. Trained with a unified spatio-temporal and physically grounded framework, RynnBrain 1.1 supports embodied perception, spatial reasoning, localization, and planning. Compared with RynnBrain 1.0, it further introduces contact-point prediction across the model family and native 3D grounding for the 2B and 9B models, yielding representations and outputs that are more directly aligned with robot manipulation. We also develop RynnBrain-VLA with a unified cross-embodiment action space and embodiment-specific masking, and deploy it on Unitree G1, Astribot-S1, and Tianji-Wuji. RynnBrain 1.1 achieves strong results on embodied cognition, localization, and 3D grounding, with the 122B-A10B model outperforming all evaluated proprietary and open-source models on VSI-Bench, MMSI, and RefSpatial-Bench. Real-robot experiments show that RynnBrain-initialized policies outperform Qwen-based and representative generalist VLAs, while joint multi-task and multi-embodiment training improves process scores and success rates over per-task training.

\vspace{-1em}
\begin{center}
    \renewcommand{\arraystretch}{1.2}
    \begin{tabular}{ll}
        \homepage  & \url{https://alibaba-damo-academy.github.io/RynnBrain} \\
        \github  & \url{https://github.com/alibaba-damo-academy/RynnBrain}\\
        \huggingface & \url{https://huggingface.co/collections/Alibaba-DAMO-Academy/rynnbrain-11} \\
    \end{tabular}
    \vspace{-1em}
\end{center}

}

\date{July 31, 2026}

\begin{document}

\thispagestyle{firstheader}
\maketitle
\pagestyle{empty}
\setcounter{tocdepth}{3}


\providecommand{\rev}[1]{{\color{red}#1}}
\providecommand{\TODO}[1]{{\color{red}\textbf{[TODO: #1]}}}

\FloatBarrier
\vspace{-0.6em}
\begin{figure*}[!h]
    \centering
    \includegraphics[width=0.92\linewidth]{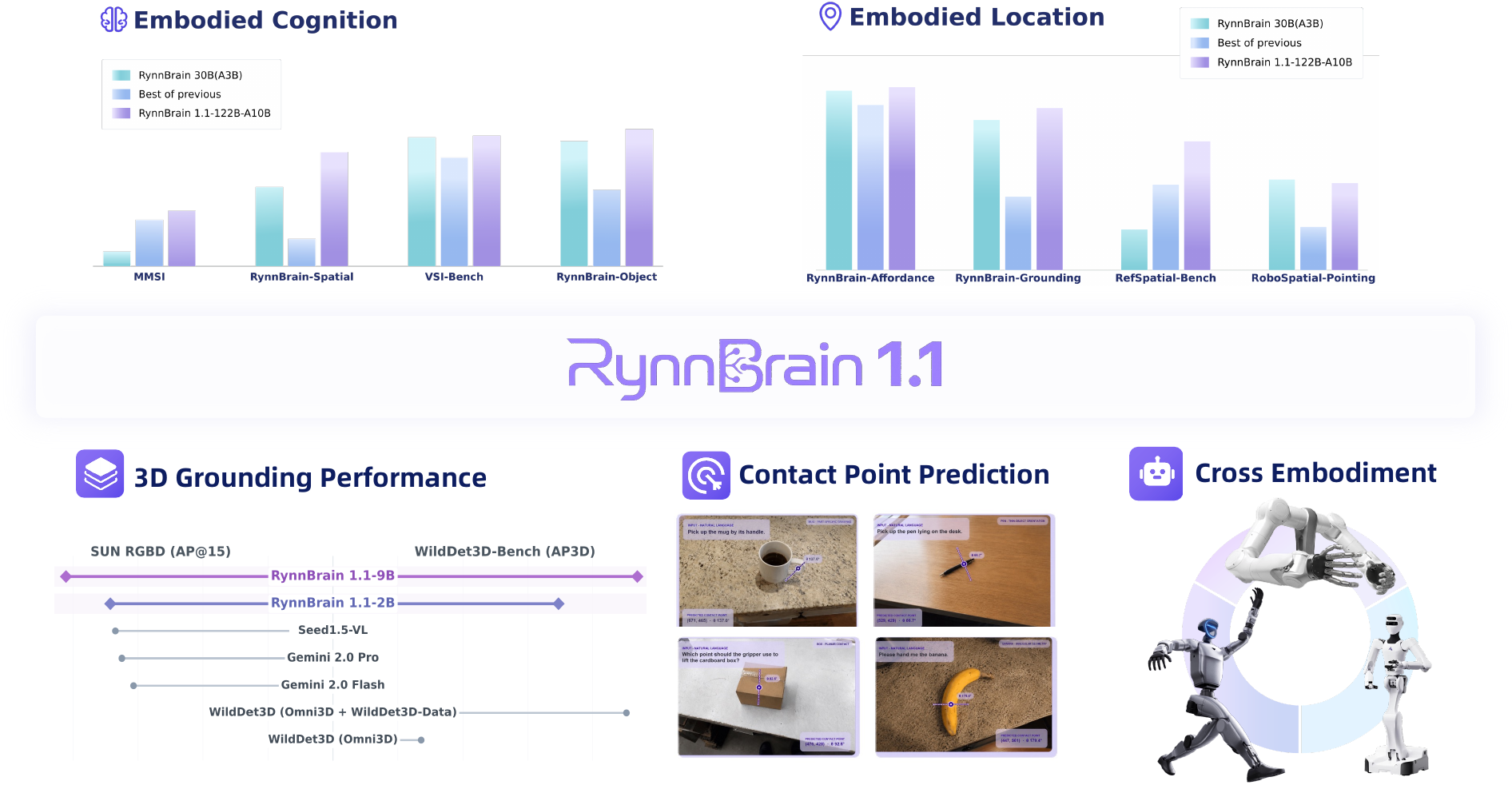}
    \caption{
    We introduce RynnBrain 1.1, a model family comprising three variants—2B, 9B, and 122B-A10B—together with the post-trained RynnBrain-VLA. RynnBrain 1.1 achieves leading performance across a broad suite of benchmarks. RynnBrain-VLA demonstrates significant promise in enabling action prediction across diverse embodiments within a unified action space.
    }
    \label{fig:main}
\end{figure*}
\FloatBarrier

\noindent

\section{Introduction} \label{sec:introduction}

Embodied intelligence requires models to go beyond visual recognition and language understanding. A robot operating in the physical world must perceive objects~\cite{ren2024grounding,ravi2024sam}, reason about spatial relations~\cite{song2025robospatial,zhou2025roborefer}, identify actionable regions~\cite{hao2025roboafford++}, and translate such understanding into executable behaviors~\cite{pi_0.5,pi0}. Although recent multimodal large language models have achieved strong performance in image and video question answering tasks~\cite{bai2025qwen3vltechnicalreport,zhang2025videollama,hou2026interlv,gemini_2.5}, real-world robotic systems impose additional requirements: they must understand 3D structure, reason across viewpoints, ground language instructions into physical locations, and support downstream manipulation policies. Therefore, the central challenge for embodied foundation models is not only whether they can understand visual content, but whether visual understanding is grounded in the physical world and, crucially, whether it can be transferred to real-robot action~\cite{team2025gemini}.

Recent embodied foundation models~\cite{MiMo-Embodied,tan2026robobrain2.5,damo2026rynnbrain,cosmos3,hyembodied} have advanced rapidly along different axes, including object and region localization, spatio-temporal understanding, and robot-oriented planning. Specifically, RynnBrain 1.0~\cite{damo2026rynnbrain} explored this direction by introducing a unified embodied multimodal model for egocentric understanding, spatial grounding, physically grounded reasoning, and planning~\cite{jia2022egotaskqa,ecbench,pei2025egothinker}, demonstrating the feasibility of combining general and robot-oriented multimodal understanding. Despite its strong performance, several key aspects were not fully explored in RynnBrain 1.0. First, how can the representations and outputs from RynnBrain be brought closer to robot manipulation? Second, how effectively can RynnBrain serve as an initialization for downstream VLA post-training?

Motivated by these two questions, we develop RynnBrain 1.1, a new family of embodied foundation models upgraded from RynnBrain 1.0~\cite{damo2026rynnbrain}. Concretely, to learn representations that are more meaningful for robot manipulation, we add two new training tasks to the embodied pretraining of RynnBrain 1.1: (1) \textbf{contact point prediction} and (2) \textbf{3D grounding}. The contact point prediction task requires the model to locate the center point and the rotation angle of the gripper for grasping. Compared to the grasp rectangle detection task in~\cite{vuong2024grasp,damo2026rynnbrain}, contact point prediction can guide the model to provide more precise (i.e., the contact point) and meaningful (i.e., the planar rotation angle of the gripper) features for the downstream robot policy. As for 3D grounding, it introduces explicit and massive 3D modeling during the training of RynnBrain 1.1. Since robots operate in three-dimensional environments~\cite{wen2024foundationpose,huang2026wilddet3d}, equipping embodied foundation models with explicit 3D modeling benefits manipulation through both richer representations and directly usable 3D outputs. In addition, to support the fine-tuning of RynnBrain 1.1 across heterogeneous robot embodiments and control interfaces, we
introduce a unified action space that organizes embodiment-specific actions into semantically aligned body-part
groups. Embodiment-specific masks are then applied to activate only the dimensions available on each robot.
This design allows data from different embodiments to be trained jointly within a single policy without forcing explicit alignment between incompatible action spaces.





Built upon Qwen3.5~\cite{qwen3.5}, RynnBrain 1.1 is available in three model scales—2B, 9B, and 122B-A10B—all trained under a unified capability definition and training framework. This consistent design enables us to systematically study how model capabilities evolve with scale.
We evaluate RynnBrain 1.1 from two complementary perspectives. First, we benchmark the base models on a broad suite of embodied multimodal tasks covering spatio-temporal understanding, spatial reasoning, localization, and 3D grounding. The results show an upward trend from 2B to 9B and 122B-A10B, while the 2B and 9B models demonstrate strong language-conditioned 3D grounding after incorporating explicit 3D supervision. Second, we evaluate how large-scale embodied pretraining helps to learn downstream VLA policy. We perform evaluations of RynnBrain-VLA on three different platforms, a Unitree G1 humanoid\footnote{\url{https://www.unitree.com/g1}}, an Astribot-S1 bimanual robot\footnote{\url{https://www.astribot.com/en/product}}, and a Tianji-Wuji dexterous-hand system\footnote{\url{https://www.tianjizn.com/products/marvin-series/}, \url{https://www.wuji.tech/zh/hand}}. Controlled evaluations on Astribot and Tianji-Wuji show that VLA policies built on our pretrained RynnBrain 1.1 outperform both the Qwen-based VLA policies and the most representative VLA generalists~\cite{pi_0.5,gr00tn1_2025}. Notably, joint multi-task and multi-embodiment training further improves the average process score and final success rate over separately fine-tuned per-task policies.

Overall, the contributions of this technical report are as follows:
\begin{itemize}
    \item We release RynnBrain 1.1, a new family of embodied foundation models spanning 2B, 9B, and 122B-A10B parameter scales. Extensive evaluations across various benchmarks demonstrate a clear upward trend in model capabilities as the parameter scale increases. Notably, our RynnBrain 1.1-122B-A10B surpasses all proprietary models as well as open-source models on VSI-Bench, MMSI, and RefSpatial-Bench.  

    \item We introduce two new features: (1) contact point prediction and (2) native 3D grounding (available for 2B and 9B-sized models) into embodied pretraining, yielding representations and outputs that are more directly aligned with robot manipulation.

    \item We develop RynnBrain-VLA, which couples a unified cross-embodiment action space with embodiment-specific masking. Beyond delivering strong real-world performance across heterogeneous embodiments, tasks, and control interfaces, it obtains further gains from joint multi-task and multi-embodiment training over separately fine-tuned per-task policies.
\end{itemize}

\section{Overview}
\subsection{Model Architecture}
\begin{figure*}
    \centering
    \includegraphics[width=\linewidth]{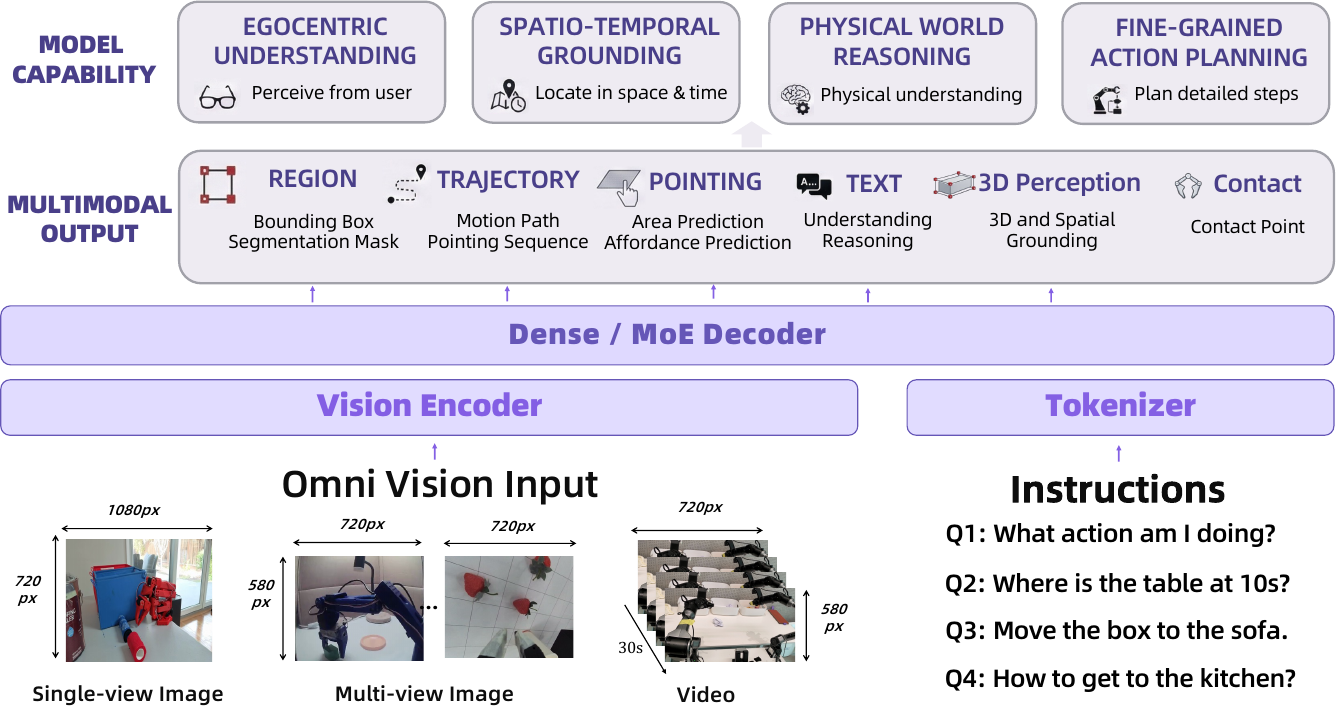}
    \caption{
\textbf{Overview of the RynnBrain 1.1 architecture}. Given language instructions and omni-visual inputs—including single-view images, multi-view images, and videos—RynnBrain produces a unified set of multimodal outputs through either a dense or a mixture-of-experts decoder. These outputs cover text, regions, trajectories, pointing, 3D perception, and contact signals, enabling egocentric understanding, spatiotemporal grounding, physically grounded reasoning, and fine-grained action planning in real-world environments.    }
    \label{fig:model framework}
\end{figure*}

An overview of the RynnBrain architecture is shown in \Cref{fig:model framework}.

RynnBrain 1.1 adopts a decoder-only vision--language architecture following the design principles of Qwen3.5~\cite{qwen3.5}. It consists of a vision encoder, a vision--language projector, and a large language model (LLM) backbone initialized from Qwen3.5 variants. In this release, we build three model scales, namely RynnBrain 1.1-2B, RynnBrain 1.1-9B, and RynnBrain 1.1-122B-A10B. All models share the same overall architecture and training formulation, enabling a systematic study of embodied scaling across model sizes. We also adopt DeepStack~\cite{meng2024deepstack} and Interleaved MRoPE~\cite{huang2025revisiting} to better integrate multimodal information and support long-context spatio-temporal modeling over images, videos, and embodied observations.

\subsection{Training Infrastructure}

As described in RynnBrain~\cite{damo2026rynnbrain}, our training pipeline is designed for large-scale embodied multimodal data with highly heterogeneous sequence lengths. We follow
the same infrastructure design, including online sequence-length-aware load balancing, per-sample loss reduction, memory-efficient training with ZeRO and gradient checkpointing, and optimized MoE execution with expert parallelism and efficient token dispatching. These
components allow us to train on mixed image, video, text, and spatial-grounding data while mitigating straggler effects caused by long-tail sequence-length distributions.

For RynnBrain 1.1, we extend this infrastructure to the new Qwen3.5-based model family,
covering 2B, 9B, and 122B scales. The same training pipeline is used across model sizes to
ensure that differences in downstream performance mainly reflect model scaling and data
updates rather than changes in infrastructure. For the 122B-A10B model, we adopt sparse MoE training with expert parallelism and communication-efficient token dispatching, following the infrastructure principles established
in RynnBrain~\cite{damo2026rynnbrain}.

\section{Pretraining}
\noindent
Similar to RynnBrain~\cite{damo2026rynnbrain}, our embodied pretraining framework is built around two key principles: (1) spatio-temporal memory and (2) physical-world grounding. Spatio-temporal memory enables the model to aggregate information from past visual observations, including objects, locations, events, and motion dynamics. Physical-world grounding, in turn, encourages the model to represent part of its understanding through explicit spatial outputs—such as objects, regions, affordances, and trajectories—rather than relying solely on unconstrained language generation. RynnBrain 1.1 advances this framework in two ways. First, we apply the same spatio-temporal embodied pretraining paradigm to three model scales—2B, 9B, and 122B-A10B—allowing us to examine how embodied perception, reasoning, localization, and planning change with model capacity. Second, as an initial step toward native 3D grounding, we add large-scale 3D-grounded data to the pretraining mixtures of RynnBrain 1.1-2B and RynnBrain 1.1-9B, enabling us to study the contribution of explicit 3D supervision to physical-space understanding at different model scales.

\subsection{Training Recipe}

Our pretraining formulation unifies temporal visual modeling and explicit spatial prediction within a single autoregressive framework. Multimodal observations are first encoded into a common token sequence, while both linguistic responses and grounded spatial targets are generated through the same prediction interface. This design equips RynnBrain with temporal memory and physically grounded representations without introducing separate task-specific decoders.

\paragraph{Unified Spatio-temporal Representation.}
A visual observation is represented as an ordered sequence of frames,
$\mathbf{V}=\{I_t\}_{t=1}^{T}$. Static images correspond to the special case $T=1$, whereas videos contain multiple temporally ordered frames with $T>1$. For video inputs, we uniformly sample frames over time and encode each frame into visual tokens. Temporal position information is then attached to the resulting tokens so that the model can distinguish frame order and associate observations across time. By casting images and videos into the same sequential form, RynnBrain 1.1 can model object persistence, motion evolution, and long-range visual dependencies using a shared architecture.

\paragraph{Physically Grounded Output Space.}
Physical grounding is incorporated directly into the model's generation vocabulary. In addition to natural-language tokens, the output sequence may contain discrete representations of bounding boxes $\mathcal{B}$, points $\mathcal{P}$, and trajectory waypoints $\mathcal{T}$. Continuous image coordinates are first mapped to integers in $[0, 1000]$ and are then predicted autoregressively in the same manner as text tokens. As a result, semantic reasoning and spatial localization are learned jointly rather than handled by separate prediction heads. The 2B, 9B, and 122B-A10B models use this shared output specification in the main scaling experiments, while differences among them arise from model capacity and their corresponding training-data mixtures.

In addition, we introduce explicit 3D modeling task into the pretraining of RynnBrain 1.1-2B and RynnBrain 1.1-9B to explore native 3D grounding. These models follow
the same instruction-following and autoregressive training paradigm as the rest of the
RynnBrain 1.1 family, while extending the supervision from image-plane grounding toward
physical-space understanding.

Specifically, the model takes as input a natural language instruction and camera intrinsics, and is trained to directly predict a 3D bounding box parameterized in the camera coordinate system---including center position $(\mathrm{cx}, \mathrm{cy}, \mathrm{cz})$, box dimensions $(\mathrm{w}, \mathrm{l}, \mathrm{h})$, and orientation $(\mathrm{pitch}, \mathrm{yaw}, \mathrm{roll})$, all in physical units (meters and radians). This 9-dimensional output format follows the design of prior works such as Seed-VL \cite{seed2026seed18modelcardgeneralized} and Qwen3-VL \cite{bai2025qwen3vltechnicalreport}, which also adopt a compact set of numerical parameters for spatially grounded predictions. To preserve the unified autoregressive framework, we discretize all continuous 3D quantities into integer tokens within a fixed range, analogous to our 2D spatial tokenization. This design enables the model to perform physically grounded reasoning from a single image, inferring not only what and where in the pixel plane, but also how objects are positioned and oriented in real-world 3D space, directly bridging visual perception and downstream robotic or navigation tasks.

\paragraph{Optimization.}
All RynnBrain 1.1 variants are optimized end-to-end with a causal autoregressive objective over sequences containing both language and spatial tokens. Given a visual input $\mathbf{V}$ and a target sequence $\mathbf{y}=\{y_i\}_{i=1}^{L}$, the model minimizes the negative log-likelihood
\begin{equation}
\mathcal{L}
=
-\sum_{i=1}^{L}
\log P\left(
y_i \mid y_{<i}, \mathbf{V}; \mathbf{\Theta}
\right).
\end{equation}
Here, $L$ denotes the target-sequence length and $\mathbf{\Theta}$ denotes the learnable model parameters. The same optimization objective is used for the entire RynnBrain 1.1 family, including the 3D-grounded 2B and 9B variants. We select scale-specific hyperparameters through preliminary experiments on representative subsets of the pretraining corpus, with the complete configurations summarized in \Cref{tab:hyperparams}.

\begin{table}[th]
    \centering
    \caption{Hyperparameters of the pretraining stage for the RynnBrain model series.}
    \label{tab:hyperparams}
    
    \renewcommand{\arraystretch}{1.05} 
    \setlength{\tabcolsep}{10pt}        
    
    \resizebox{\textwidth}{!}{
    \begin{tabular}{lccc}
        \toprule
        \textbf{Parameter} & \textbf{RynnBrain 1.1-2B} & \textbf{RynnBrain 1.1-9B} & \textbf{RynnBrain 1.1-122B-A10B} \\
        \midrule
        Base Model           & Qwen3.5-2B   & Qwen3.5-9B   & Qwen3.5-122B-A10B \\
        Optimizer            & AdamW        & AdamW        & AdamW \\
        Learning Rate        & $5\times10^{-6}$ & $2\times10^{-6}$ & $2\times10^{-6}$ \\
        Learning Rate Vision & $1\times10^{-6}$ & $2\times10^{-6}$ & $2\times10^{-6}$ \\
        Global Batch Size    & 512          & 1024         & 1024 \\
        Warmup Ratio         & 0.03         & 0.03         & 0.03 \\
        \bottomrule
    \end{tabular}
    }
\end{table}


\subsection{Pretraining Data}
\label{sec:pretraindata}

\newcommand{\newdata}[1]{\color{rynn}{#1}}

\begin{table}[t]
    \centering
    \caption{Pretraining data mixture for RynnBrain 1.1. New or substantially expanded data sources compared with RynnBrain 1.0 are highlighted in \newdata{blue}.}
    \label{tab:data-mixture}

    \begin{tabularx}{\textwidth}{
        >{\raggedright\arraybackslash}p{2.2cm}|
        >{\raggedright\arraybackslash}p{3.6cm}|
        >{\raggedright\arraybackslash}X
    }
        \hline
        \textbf{Category} & \textbf{Sub-Task} & \textbf{Data Sources} \\
        \hline

        \multirow{1}{*}{\textbf{General MLLM}}
        & General
        & LLaVA-OV-SI~\cite{llava-OV},
        LLaVA-Video~\cite{llava-video},
        ShareGPT-4o-video~\cite{sharegpt4video},
        VideoGPT-plus~\cite{videogpt+},
        FineVideo~\cite{FineVideo},
        CinePile~\cite{cinepile},
        ActivityNet~\cite{activitynet},
        YouCook2~\cite{youcook2},
        LLaVA-SFT~\cite{llava},
        \newdata{VL3~\cite{zhang2025videollama}},
        \newdata{RynnBrain-Thinking}\\
        \hline

        \multirow{5}{*}{\textbf{Cognition}}
        & Object Understanding
        & RynnBrain-Object,
        RefCOCO~\cite{yu2016modeling},
        Google Refexp~\cite{mao2016generation},
        Osprey-724K~\cite{yuan2024osprey},
        DAM~\cite{lian2025describe},
        VideoRefer-700k~\cite{yuan2025videorefer} \\

        & Spatial Understanding
        & SenseNova-SI-800K~\cite{cai2025scaling},
        VSI-590K~\cite{yang2025cambrian},
        VLM-3R~\cite{fan2025vlm},
        RynnBrain-Spatial \\

        & Counting
        & RynnBrain-Counting,
        Molmo2~\cite{clark2026molmo2}
        \\

        & OCR
        & RynnBrain-OCR,
        \newdata{Llama-Nemotron-VLM OCR}~\cite{llama_nemotron_vlm_data} \\

        & Egocentric Task Understanding
        & EgoRe-5M~\cite{pei2025egothinker},
        EgoTaskQA~\cite{jia2022egotaskqa},
        Env-QA~\cite{Gao_2021_ICCV},
        QAEgo4D~\cite{grauman2022ego4d},
        RoboVQA~\cite{sermanet2024robovqa},
        Robo2VLM~\cite{chen2025robo2vlm},
        ShareRobot~\cite{ji2025robobrain}
        \\
        \hline

        \multirow{7}{*}{\textbf{Localization}}
        & Object Localization
        & ADE20K~\cite{ade20k},
        COCOStuff~\cite{cocostuff},
        Mapillary~\cite{mapillary},
        PACO-LVIS~\cite{paco},
        PASCAL-Part~\cite{pascal-part},
        VG~\cite{krishna2017visual},
        RoboAfford-Object~\cite{hao2025roboafford++},
        RynnBrain-Grounding \\

        & Area Localization
        & RefSpatial~\cite{zhou2025roborefer},
        RoboAfford-Area~\cite{hao2025roboafford++},
        Molmo2~\cite{clark2026molmo2},
        RynnBrain-Area,
        \newdata{RoboRefer~\cite{zhou2025roborefer}},
         \\

        & Affordance Localization
        & RynnBrain-Affordance,
        RoboAfford-Affordance~\cite{hao2025roboafford++}
        \\

        & Trajectory Prediction
        & RynnBrain-Trajectory,
        FSD~\cite{yuan2025seeingdoingbridgingreasoning},
       \newdata{MolmoAct~\cite{lee2025molmoact}} 
       \\

        & 3D Grounding
        & \newdata{WildDet3D Essential~\cite{huang2026wilddet3d}},
        \newdata{WildDet3D Synthetic~\cite{huang2026wilddet3d}},
        \newdata{FoundationPose~\cite{wen2024foundationpose}} \\

        & Grasp / Contact Point Prediction
        &
        \newdata{Grasp-Anything-Visible~\cite{vuong2024grasp}},
        \newdata{GraspClutter6D~\cite{back2025graspclutter6d}},
        \newdata{GraspNet-1B~\cite{fang2020graspnet}},
        \newdata{Jacquard V2~\cite{li2024jacquard}},
        \newdata{GraspFactory~\cite{srinivas2025graspfactory}} \\
        \hline

        \multirow{1}{*}{\textbf{Planning}}
        & Manipulation
        & AgibotWorld~\cite{contributors2024agibotworldrepo},
        Open X-Embodiment~\cite{open_x_embodiment_rt_x_2023},
        RynnBrain-Planning,
        \newdata{Galaxea-G0}~\cite{galaxea_g0}
        \\
        \hline
    \end{tabularx}
\end{table}

RynnBrain 1.1 follows the capability-oriented data organization of RynnBrain~\cite{damo2026rynnbrain}. As shown in \Cref{tab:data-mixture}, we organize the pretraining mixture into general multimodal data, cognition data, spatio-temporal localization data, 3D-grounded data, robot-oriented perception data, and planning data. For data types already introduced in RynnBrain 1.0, we follow the same instruction-following format and data processing protocol. Specifically, visual inputs are represented as images or videos, and model outputs are converted into mixed sequences of text and spatial tokens, including boxes, points, trajectories, and grasp-related annotations. In this section, we focus on the data updates introduced in RynnBrain 1.1.

\paragraph{General MLLM Data.}
To preserve broad multimodal understanding, we continue to use a general-purpose image--video--text corpus covering image understanding, video understanding, visual question answering, temporal reasoning, and instruction following. The mixture includes commonly used MLLM datasets such as LLaVA-OV-SI~\cite{llava-OV}, LLaVA-Video~\cite{llava-video}, ShareGPT-4o-video~\cite{sharegpt4video}, VideoGPT-plus~\cite{videogpt+}, FineVideo~\cite{FineVideo}, CinePile~\cite{cinepile}, ActivityNet~\cite{activitynet}, YouCook2~\cite{youcook2}, and LLaVA-SFT~\cite{llava}. In RynnBrain 1.1, we further incorporate self-collected RynnBrain-Thinking data. RynnBrain-Thinking is included as part of the general mixture to better align the model with the Qwen3.5-Thinking style. All data are converted into the unified instruction-following format before being mixed for pretraining.

\paragraph{Multi-dimensional Cognition Data.}
For embodied cognition, RynnBrain 1.1 follows the task taxonomy of RynnBrain~\cite{damo2026rynnbrain}, covering object understanding, spatial understanding, counting, OCR, and egocentric task understanding. Object understanding data include RynnBrain-Object, RefCOCO~\cite{yu2016modeling}, Google Refexp~\cite{mao2016generation}, Osprey-724K~\cite{yuan2024osprey}, DAM~\cite{lian2025describe}, and VideoRefer-700K~\cite{yuan2025videorefer}, which provide object-centric supervision for fine-grained recognition and reasoning. Spatial understanding data include SenseNova-SI-800K~\cite{cai2025scaling}, VSI-590K~\cite{yang2025cambrian}, VLM-3R~\cite{fan2025vlm}, and RynnBrain-Spatial, supporting reasoning over spatial relations, viewpoints, and scene structure. For counting and OCR, we use RynnBrain-Counting, Molmo2~\cite{clark2026molmo2}, RynnBrain-OCR, and the newly added Llama-Nemotron-VLM OCR data. For egocentric task understanding, we use EgoRe-5M~\cite{pei2025egothinker}, EgoTaskQA~\cite{jia2022egotaskqa}, Env-QA~\cite{Gao_2021_ICCV}, QAEgo4D~\cite{grauman2022ego4d}, RoboVQA~\cite{sermanet2024robovqa}, Robo2VLM~\cite{chen2025robo2vlm}, and ShareRobot~\cite{ji2025robobrain}. These datasets are converted following the RynnBrain instruction format to provide broad supervision for object-centric, spatial, textual, and first-person task understanding.

\paragraph{Spatio-temporal Localization Data.} RynnBrain 1.1 inherits the localization formulation of RynnBrain 1.0, where spatial outputs are represented using normalized coordinate tokens. The localization mixture covers object localization, area localization, affordance localization, trajectory prediction, and grasp/contact-related prediction. Object localization uses ADE20K~\cite{ade20k}, COCOStuff~\cite{cocostuff}, Mapillary~\cite{mapillary}, PACO-LVIS~\cite{paco}, PASCAL-Part~\cite{pascal-part}, VG~\cite{krishna2017visual}, RoboAfford-Object~\cite{hao2025roboafford++}, and RynnBrain-Grounding. Area localization uses RefSpatial~\cite{zhou2025roborefer}, RoboAfford-Area~\cite{hao2025roboafford++}, Molmo2~\cite{clark2026molmo2}, RynnBrain-Area, and the newly added RoboRefer. Affordance localization uses RynnBrain-Affordance and RoboAfford-Affordance~\cite{hao2025roboafford++}. Trajectory prediction uses RynnBrain-Trajectory, FSD~\cite{yuan2025seeingdoingbridgingreasoning}, and the newly added MolmoAct. For these data sources, we follow the RynnBrain processing protocol and convert different annotations into a shared output space consisting of boxes, points, and trajectories.

\paragraph{3D-grounded Data.} 
To explore native 3D grounding, we conduct a 3D-grounded preview training in the 2B and 9B setting by mixing explicit 3D annotations into the pretraining data.  Given an image, a natural language instruction, and camera intrinsics, the model is trained to predict a 3D bounding box in the camera coordinate system. The box is parameterized by its center position $(\mathrm{cx}, \mathrm{cy}, \mathrm{cz})$, dimensions $(\mathrm{w}, \mathrm{l}, \mathrm{h})$, and orientation $(\mathrm{pitch}, \mathrm{yaw}, \mathrm{roll})$, represented in physical units such as meters and radians. 

To support this at scale, we construct a multi-source corpus combining large-scale automatically lifted annotations with high-fidelity synthetic data. For broad category and scene coverage, we adopt \textbf{WildDet3D-Data}~\cite{huang2026wilddet3d}, a recently proposed open-vocabulary 3D detection dataset that lifts 2D annotations from COCO, LVIS, Objects365, and V3Det into 3D via a multi-model candidate generation and selection pipeline. We use its two curated subsets after additional filtering: \textit{Essential} (102,979 images, 374K manually verified annotations after filtering out geometrically implausible candidates and small objects, 12,064 categories) and \textit{Synthetic} (896,004 images, 888K automatically selected annotations, 11,896 categories, filtered by a fine-tuned Molmo2 scoring six perceptual criteria with a retention threshold >10), offering unprecedented category diversity but inheriting residual noise from monocular lifting. For each 3D annotation, we leverage Qwen-VL to generate an open-ended referring query based on the corresponding 2D bounding box and category label, incorporating spatial descriptions (e.g., relative positions such as "left," "right," "front," or "behind") and attribute modifiers (e.g., color, size, or material cues) to make the referring expression more discriminative and precise for grounding. To complement this with physically exact supervision, we additionally incorporate \textbf{FoundationPose}~\cite{wen2024foundationpose}, a synthetic dataset derived from the Google Scanned Objects subset under controlled camera trajectories, which provides 1,849K images with noise-free 3D bounding boxes computed directly from ground-truth object-to-camera transformations, with each asset assigned a natural-language category via Qwen-VL to bridge to real-world semantics. Though its category coverage is limited to 928 instances, its metric-scale accuracy and zero-depth-ambiguity geometry serve as a strong regularizer for depth and orientation prediction, teaching correct inductive biases for camera-to-world projection when mixed with the diverse but noisier WildDet3D annotations.

\paragraph{Contact Point Data. }
RynnBrain 1.0 represented planar grasps using the four corners of an oriented rectangle. We found this detector-style representation poorly aligned with action grounding in a general embodied model. 
A target object typically admits a set of functionally valid grasps rather than a unique rectangle. Moreover, unlike an object bounding box, a grasp rectangle has no canonical image-space extent: its width, height, aspect ratio, and boundary placement depend on gripper geometry and dataset-specific annotation conventions.
Consequently, IoU against a single reference rectangle may penalize functionally equivalent grasps and is not necessarily correlated with execution success.
The four-corner representation also introduces redundant extent variables and geometric consistency constraints into autoregressive generation.
RynnBrain 1.1 therefore adopts a compact contact-centered representation $(a=(p,\theta))$, where $(p=(x,y))$ denotes the center of the annotated grasp configuration and $(\theta)$ is the in-plane angle of the line connecting the gripper fingers.
The coordinates are normalized to [0,1000], and each target is serialized as <grasp pose> (x, y), $\theta$ </grasp pose>.
This representation factors out annotation- and embodiment-dependent rectangle extents while preserving an action-relevant spatial anchor and orientation cue.

A substantial portion of our source data consists of simulation-native, multidimensional grasp annotations rather than directly usable image–text pairs. 
We therefore develop a view-conditioned rendering and projection pipeline to convert these annotations into reliable 2D supervision. For each object–grasp configuration, we select a camera pose subject to multiple observability constraints: the target object must remain fully inside the image, occupy an appropriate range of image scales, and avoid severe truncation or occlusion; meanwhile, the projected grasp center and orientation must both remain visible and spatially distinguishable.
Views in which the object is excessively near, distant, truncated, or visually ambiguous, or in which the grasp configuration is not observable, are discarded. The retained views are rendered as RGB images, after which the multidimensional grasp annotations are projected into the common $((p,\theta))$ representation. This procedure is essential because unconstrained camera sampling can produce geometrically valid annotations that cannot be inferred from the rendered image.

After rendering, projection, and filtering, the resulting corpus contains 2.6M training samples from Grasp-Anything~\cite{vuong2024grasp}, Jacquard V2~\cite{li2024jacquard}, GraspFactory~\cite{srinivas2025graspfactory}, GraspNet-1B~\cite{fang2020graspnet}, and GraspClutter6D~\cite{back2025graspclutter6d}. We treat $((p,\theta))$ as an action-grounding interface rather than a complete executable grasp pose; depth, approach direction, gripper aperture, collision constraints, and robot kinematics are resolved by the downstream manipulation policy.

\paragraph{Physics-aware Planning Data.} For manipulation planning, RynnBrain 1.1 follows the structured planning format of RynnBrain~\cite{damo2026rynnbrain}, where high-level task instructions are paired with grounded sub-task descriptions involving objects, areas, and affordances. The planning mixture includes AgibotWorld~\cite{contributors2024agibotworldrepo}, Open X-Embodiment~\cite{open_x_embodiment_rt_x_2023}, RynnBrain-Planning, and the newly added Galaxea-G0. For existing grounded planning data, we follow the same annotation format and conversion protocol as RynnBrain 1.0. For Galaxea-G0, the original annotations provide sub-task-level labels but do not contain explicit high-level task instructions. We therefore construct the corresponding global task descriptions using Gemini 3.1 Pro, conditioned on the annotation IDs and the sequence of annotated sub-tasks. The resulting data are then converted into the same grounded planning format as the rest of the planning mixture. 

\providecommand{\rev}[1]{{\color{red}#1}}
\providecommand{\TODO}[1]{{\color{red}\textbf{[TODO: #1]}}}
\section{Post-training for Vision-Language-Action Model}
\label{sec:vla}

\noindent
We use downstream VLA post-training as both a practical capability and a probe of base-model quality. This section describes the VLA \emph{method}: the policy architecture (\Cref{sec:vla_arch}), the unified action space design (\Cref{sec:unified_action_space}), the cross-embodiment / cross-driver deployment framework (\Cref{sec:vla_deploy}), Real-Time Chunking for smooth low-latency control (\Cref{sec:vla_rtc}), and the fine-tuning setup shared by all backbones (\Cref{sec:vla_finetune}). The corresponding real-robot results are reported in \Cref{sec:realrobot}.

\begin{figure*}
    \centering
    \includegraphics[width=\linewidth]{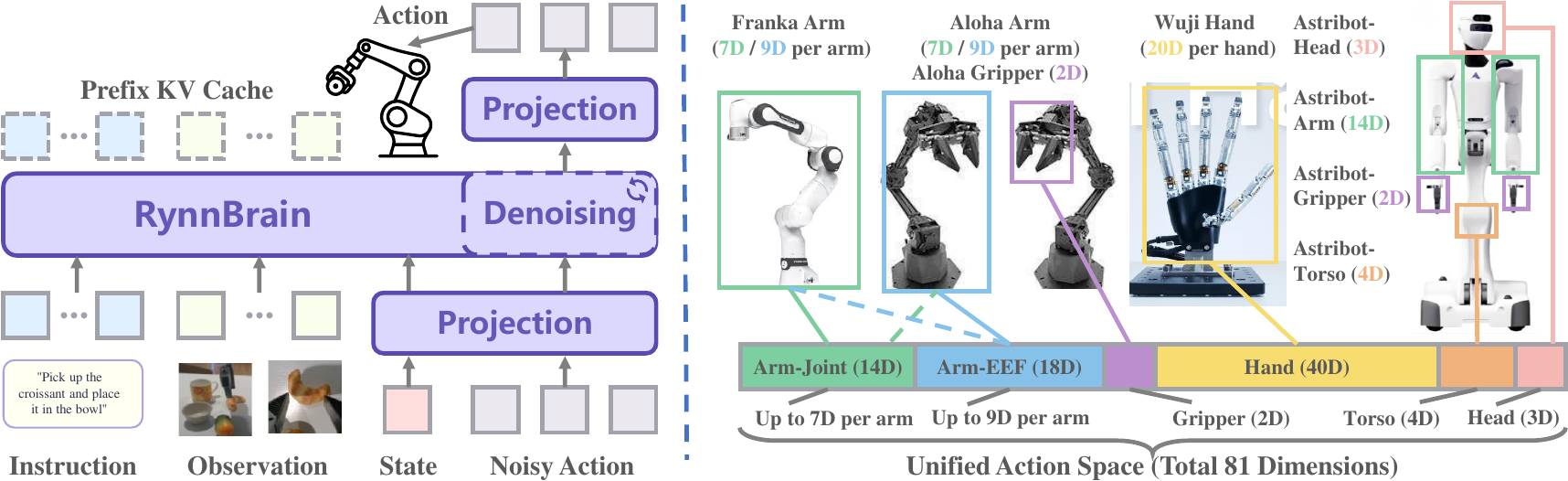}
    \caption{\textbf{RynnBrain-VLA architecture and unified action space.}
    \textit{(Left)} RynnBrain-VLA adopts a flow-matching framework in which the RynnBrain base model serves as a single-stream Diffusion
    Transformer (DiT). The model takes a packed sequence of language instruction, multi-camera observations, robot state, and noisy action chunks
    as input, and iteratively denoises to predict an action chunk. The instruction token prefix is cached via KV cache to reduce inference latency.
    \textit{(Right)} Actions from heterogeneous robot embodiments are represented within a shared 81-dimensional action space partitioned into
    semantically aligned body-part groups: Arm-Joint (14D, up to 7D per arm), Arm-EEF (18D, up to 9D per arm), Gripper (2D), Hand (40D), Torso (4D), and Head (3D).
    Each embodiment activates only its physically available dimensions via an embodiment-specific mask, enabling joint training across robots with incompatible low-level action spaces.
    In our evaluated embodiments: Unitree G1 activates Hand (14D, 7D per hand) within the unified space alongside a separately predicted 64D SONIC~\cite{luo2025sonic} latent token;
    Tianji-Wuji (Tianji Arm + Wuji Hand) activates Arm-Joint (14D) and Hand (40D);
    Astribot S1 activates Arm-Joint (14D), Gripper (2D), Head (3D) and Torso (4D).
    The robots shown in the figure are illustrative examples of the masking mechanism.
}
    \label{fig:vla_model}
\end{figure*}

\subsection{Model Architecture}
\label{sec:vla_arch}
As shown in \Cref{fig:vla_model}, the model architecture of \textbf{RynnBrain-VLA} in RynnBrain-1.1 largely follows that in RynnBrain-1.0~\cite{damo2026rynnbrain}, where the VLM backbone itself serves as a single-stream diffusion transformer and processes dense interaction among vision, language, and action in each layer. A flow matching framework is adopted to predict an action chunk~\cite{pi0} at each step.
To preserve the VLM's inherent instruction-following capabilities, we utilize its native conversation format for organizing the input:


\fcolorbox{black}{black!5}{
\begin{minipage}{0.975\textwidth}
<|im\_start|>user \\
INSTRUCTION: \\
Pour the wine from the bottle into the wine glass and place the glass on the table. \\
OBSERVATION: \\
<camera\_1><camera\_2><camera\_3> \\
STATE: \\
<state> \\
What action should the robot take ?<|im\_end|> \\
<|im\_start|>assistant \\
<action>
\end{minipage}
}

Following $\pi_0$~\cite{pi0}, actions are positioned at the end of the sequence to enable the KV cache during inference.

\subsection{Unified Cross-Embodiment Action Space}
\label{sec:unified_action_space}

As illustrated in \Cref{fig:vla_model}, RynnBrain-VLA represents actions of heterogeneous robot embodiments within a shared high-dimensional action space. Before training, inspired by~\cite{liu2025rdt}, we partition the unified action vector into semantically aligned groups corresponding to different body parts and control modes, including left and right arms, parallel grippers, dexterous hands, torso, and head control. Each embodiment activates only the action groups that are physically available on that robot, while all remaining dimensions are suppressed by an embodiment-specific mask.

The action loss is computed only on the active dimensions. For the Unitree G1 humanoid, the policy activates 14 dimensions (out of 40D) of the Hand group within the unified action space. These 14 dimensions correspond to the three-fingered dexterous hands of G1 with 7D per hand. In addition, the policy predicts a separate 64-dimensional SONIC~\cite{luo2025sonic} latent motion tokens that are not part of the unified action space. The 14-dimensional hand prediction and the 64-dimensional SONIC token are concatenated into a 78-dimensional representation and then fed to the SONIC whole-body controller to calculate target joint positions for the Unitree G1. For Tianji-Wuji, the policy activates the Arm-Joint group (14D, 7D per arm) and the Hand group (40D, 20D per hand), resulting in 54 active action dimensions.  Astribot activates the Arm-Joint (14D, two arms), Gripper (2D), Head (3D), and Torso (4D), while masking the dexterous-hand joints.

This formulation allows trajectories from different robots to be mixed within the same training batch and optimized by a single VLA policy. The shared action components, such as left- and right-arm joint motion
, receive supervision from multiple embodiments, whereas the embodiment-specific components are updated only by the corresponding data. Consequently, robot-independent manipulation features—including goal-directed arm motion, grasping, bimanual coordination, and task-level interaction patterns—can be shared across embodiments without requiring explicit alignment of their heterogeneous low-level action spaces. The unified action space not only serves as a common output interface, but also provides a mechanism for cross-embodiment knowledge sharing while preserving embodiment-specific control.
\subsection{Cross-Embodiment and Cross-Driver Deployment}
\label{sec:vla_deploy}
Real-world deployment requires a single policy interface to drive robots with very different configurations 
and low-level controllers. We therefore design a cross-embodiment, cross-driver framework that decouples the VLA policy from embodiment-specific execution, enabling the \emph{same} RynnBrain-VLA recipe to be adapted to heterogeneous robots.
The framework has three components:
\begin{itemize}
    \item \textbf{Model layer.} Runs the VLA policy to produce a 32-step action chunk in the standard action space.
    \item \textbf{Control layer.} Two independent loops: a low-frequency loop (30 Hz) coordinating inference timing, the user interface, and action source switching; and a high-frequency loop (200 Hz) that interpolates the model's action chunk to the target control frequency and sends commands to the robot. The two loops communicate via shared memory.
    \item \textbf{Embodiment layer.} Dispatches standard-format actions (left/right arm, gripper, hand, torso) to the corresponding robot actuators and formats the robot's current state back for the model layer. Multi-camera images are packaged as a dictionary whose keys match the naming convention used during training.
\end{itemize}
This design allows adding a new embodiment by only implementing the embodiment layer, without modifying the model or control layers. We instantiate this framework on three heterogeneous embodiments---the Unitree G1 humanoid, the Astribot-S1 bimanual manipulator, and the Tianji-Wuji dexterous-hand system---covering distinct action spaces, degrees of freedom, and end-effectors.

\subsection{Real-Time Chunking (RTC)}
\label{sec:vla_rtc}
We adopt \textbf{Real-Time Chunking (RTC)}~\cite{black2026real}. The model predicts a 32-step action chunk but does not wait for the entire chunk to be executed: a new inference is triggered every 5 steps while the current chunk is still running. The unexecuted portion of the previous chunk is then used to guide the denoising of the new chunk, following the action-guidance formulation of~\citet{black2026real} (Sec.~3.1) with strength $\beta = 10.0$. Each step in the previous chunk is assigned a guidance weight reflecting whether it will be consumed before the new chunk arrives: (i) the initial 5 steps and the steps consumed during inference (estimated as the sliding average of the past ten inference times) receive weight $= 1$, as they are guaranteed to be executed before the new chunk is ready; (ii) a transition region where the weight decreases smoothly from 1 to 0; and (iii) the tail of the chunk, which receives weight $= 0$ with no consistency enforced. This ensures smooth chunk-boundary continuity while allowing the model to freely adapt later predictions to new observations.

\subsection{Fine-tuning Setup}
\label{sec:vla_finetune}

We collect teleoperation demonstrations on the three target embodiments described above,
covering a range of pick-and-place and manipulation tasks with different objects and
interaction requirements. Each demonstration contains synchronized visual observations,
robot states, and action trajectories, together with a task instruction specifying the target
object or placement location.

For the controlled comparison between the Qwen-based VLA and RynnBrain-VLA in
\Cref{sec:realrobot}, we use the same demonstration data, VLA post-training pipeline, and
optimization settings. Both policies are fine-tuned for 60k steps with a learning rate of
$2\times10^{-5}$ and a batch size of 32. All input images are proportionally resized such
that the short side is 384 pixels. This setup reduces differences introduced by downstream
training and allows us to examine how the choice of base model affects the resulting
real-robot policy. The fine-tuned policies are evaluated quantitatively in
\Cref{sec:realrobot}.
\providecommand{\rev}[1]{{\color{red}#1}}
\providecommand{\TODO}[1]{{\color{red}\textbf{[TODO: #1]}}}
\section{Evaluation}
\label{sec:experiment}

 We organize our evaluation into several parts. We first assess the base models on embodied cognition and localization benchmarks across three model scales. At the 2B scale, we compare RynnBrain 1.1 with RynnBrain 1.0, Cosmos 3-Edge~\cite{cosmos3}, SenseNova-SI-1.1-InternVL3-2B~\cite{sensenova-si}, Cosmos-Reason2-2B~\cite{azzolini2025cosmos}, and Qwen3.5-2B~\cite{qwen3.5}. At the 4B–9B scale, we compare against RynnBrain 1.0, Cosmos 3-Nano~\cite{cosmos3}, RoboBrain-2.5-4B~\cite{tan2026robobrain2.5}, Thinker-4B~\cite{UBTECH_Thinker_short_report}, Molmo2-ER-5B~\cite{clark2026molmo2}, Pelican-VL-7B~\cite{pelican-VL}, MiMo-Embodied-7B~\cite{MiMo-Embodied}, Cosmos-Reason2-8B~\cite{azzolini2025cosmos}, SenseNova-SI1.5-8B~\cite{sensenova-si}, and Qwen3.5~\cite{qwen3.5}. At the largest scale, we compare with Cosmos 3-Super, RoboBrain 2.0-32B, Cosmos-Reason2-32B, Pelican-VL-72B~\cite{pelican-VL}, Qwen3.5-122B-A10B~\cite{qwen3.5}, Gemini 3 Pro~\cite{gemini}, GPT 5.4~\cite{GPT-5.4}, Claude Sonnet 4.6~\cite{Claude-Sonnet-4.6}, Gemini Robotics-ER 1.5~\cite{team2025gemini}, and HY-Embodied 0.5~\cite{hyembodied}. We then analyze scaling against the corresponding Qwen3.5 models, evaluate native 3D grounding and contact-point prediction, and finally examine RynnBrain-VLA through controlled real-robot evaluations across three embodiments
 
 \noindent 
\begin{table}[t]
\caption{Comparison with lightweight (2B-scale) models on embodied cognition and localization benchmarks. \textnormal{* denotes results from our reproduction.}}
\centering
\setlength{\tabcolsep}{10.0pt}
\renewcommand{\arraystretch}{1.3}

\begin{adjustbox}{width=\textwidth}
\begin{tabular}{
p{3.5cm}
!{\color{rynn}\vrule width 1.2pt}
>{\columncolor{rynn!7}\fontsize{11.5pt}{13pt}\selectfont}c
!{\color{rynn}\vrule width 1.2pt}
c c c c c
}
\toprule
\multirow{2}{*}{\textbf{Benchmark}}
& \textbf{RynnBrain 1.1}
& \textbf{RynnBrain}
& \makecell{\textbf{Cosmos 3}\\\textbf{-Edge}}
& \makecell{\textbf{SenseNova-SI-1.1}\\\textbf{InternVL3}}
& \makecell{\textbf{Cosmos}\\\textbf{-Reason2}}
& \textbf{Qwen3.5}
\\
& \textbf{2B}
& \textbf{2B}
& \textbf{2B}
& \textbf{2B}
& \textbf{2B}
& \textbf{2B}
\\
\midrule

VSI-Bench
& \textbf{72.9} & 70.5 & 59.2 & 63.7* & 46.3* & 44.1 \\
MMSI
& \textbf{40.5} & 34.1 & 32.3 & 34.2 & 27.8* & 28.3 \\
ERQA
& \textbf{42.3} & 42.3 & 42.0 & 28.7* & 37.8 & 33.0 \\
RoboSpatial-Pointing
& 59.6 & \textbf{65.6} & - & 0.0* & 11.5* & - \\
RoboSpatial-VQA
& 60.4 & 64.9 & - & 24.8* & \textbf{65.3*} & - \\
RoboSpatial-Overall
& 62.1 & \textbf{65.2} & - & 16.1* & 47.4* & 41.5 \\
MindCube
& \textbf{61.7} & 50.1 & - & 41.8 & 33.7* & 37.5 \\
EmbSpatial
& 73.9 & \textbf{79.6} & - & 63.0* & 65.6* & 67.7* \\
RynnBrain-Object
& \textbf{74.6} & 70.7 & 24.0 & 20.7 & 25.8 & 37.1* \\
RynnBrain-Spatial
& \textbf{63.8} & 57.2 & 22.6 & 13.0 & 24.3 & 15.6* \\

RefSpatial-Bench
& \textbf{58.5} & 52.7 & 48.4 & 5.6* & 23.6* & 30.0 \\
RynnBrain-Grounding
& \textbf{84.1} & 79.1 & 51.6 & 12.6 & 14.5 & 24.2* \\
RynnBrain-Area
& \textbf{57.2} & 54.6 & 39.1 & 8.6 & 22.1 & 8.1* \\
RynnBrain-Affordance
& \textbf{90.2} & 89.4 & 77.6 & 57.8 & 55.5 & 80.4* \\
RynnBrain-Trajectory
& \textbf{68.8} & 66.6 & 58.7 & 49.0 & 43.7 & 28.3* \\
\bottomrule
\end{tabular}
\end{adjustbox}

\label{tab:rynnbrain-2b}
\end{table}

\begin{table}[t]
\caption{Comparison with 4B--9B scale models on embodied cognition and localization benchmarks. \textnormal{* denotes results from our reproduction.}}
\centering
\setlength{\tabcolsep}{1.7pt}
\renewcommand{\arraystretch}{1.5}

\begin{adjustbox}{width=\textwidth}
\begin{tabular}{
p{3.5cm}
!{\color{rynn}\vrule width 1.2pt}
>{\columncolor{rynn!7}\fontsize{11.0pt}{12.8pt}\selectfont\hspace{3pt}}c<{\hspace{4pt}}
@{\color{rynn}\vrule width 1.2pt}
>{\hspace{4pt}}c<{\hspace{4pt}}
c c c c c c c c c
}
\toprule
\multirow{2}{*}{\textbf{Benchmark}}
& \textbf{RynnBrain 1.1}
& \textbf{RynnBrain}
& \makecell{\textbf{Cosmos 3}\\\textbf{-Nano}}
& \makecell{\textbf{RoboBrain-2.5}}
& \textbf{Thinker}
& \makecell{\textbf{Molmo2-ER}}
& \textbf{Pelican-VL}
& \makecell{\textbf{MiMo-}\\\textbf{Embodied}}
& \makecell{\textbf{Cosmos}\\\textbf{-Reason2}}
& \makecell{\textbf{SenseNova}\\\textbf{-SI1.5}}
& \textbf{Qwen3.5}
\\
& \textbf{9B}
& \textbf{8B}
& \textbf{8B}
& \textbf{4B}
& \textbf{4B}
& \textbf{5B}
& \textbf{7B}
& \textbf{7B}
& \textbf{8B}
& \textbf{8B}
& \textbf{9B}
\\
\midrule

VSI-Bench
& \textbf{74.9} & 70.9 & 54.9 & 41.7 & 65.4 & 74.5 & 52.8 & 48.5 & 53.7* & 67.3* & 61.0 \\
MMSI
& \textbf{47.0} & 39.6 & 36.2 & 20.5 & 27.7* & 43.8 & 26.2* & 30.2* & 31.3* & 38.3* & 13.4 \\
ERQA
& \textbf{47.5} & 46.8 & 46.0 & 43.3 & 42.5* & 46.8 & 39.8* & 46.8 & 46.0* & 40.0* & 41.5* \\
RoboSpatial-Pointing
& \textbf{69.7} & 60.1 & 33.0* & - & - & 32.0 & - & - & 24.6 & 29.3* & 48.7 \\
RoboSpatial-VQA
& 68.0 & \textbf{80.0} & 77.6* & - & - & 73.4 & - & - & 76.3 & 61.4* & 57.0 \\
RoboSpatial-Overall
& 69.1 & \textbf{73.1} & 62.1* & 62.3 & 70.8 & 59.0* & 57.5 & 61.8 & 59.0* & 40.1* & 54.1* \\
MindCube
& 86.9 & 56.6 & 40.5* & 26.9 & 39.0* & 57.0 & 33.7* & 43.1* & 43.8* & \textbf{92.1} & 33.1 \\
EmbSpatial
& 81.9 & 80.4 & 81.4* & 76.9 & 80.2 & 78.8 & 76.6 & 76.2 & 77.0* & 80.3 & \textbf{83.0} \\
RynnBrain-Object
& \textbf{75.9} & 71.2 & 39.8 & 36.4 & 26.5 & 17.1 & 30.8 & 39.0 & 37.2 & 16.8 & 45.6 \\
RynnBrain-Spatial
& \textbf{67.9} & 59.9 & 26.1 & 32.6 & 17.3 & 19.7 & 20.5 & 28.3 & 31.4 & 19.9 & 28.4 \\

RefSpatial-Bench
& \textbf{67.2} & 59.2 & 53.1 & 56.0 & 61.0 & 52.5 & 22.3 & 48.0 & 33.1* & 8.6* & 37.3 \\
RynnBrain-Grounding
& \textbf{85.4} & 81.6 & 72.9 & 5.4 & 2.4 & 4.2 & 3.5 & 49.8 & 60.0 & 44.7 & 33.0 \\
RynnBrain-Area
& \textbf{59.6} & 56.2 & 52.0 & 42.4 & 33.8 & 30.6 & 46.5 & 49.4 & 37.6 & 11.3 & 32.9 \\
RynnBrain-Affordance
& \textbf{90.7} & 90.4 & 85.1 & 69.1 & 59.9 & 81.7 & 81.4 & 84.4 & 83.9 & 64.9 & 60.5 \\
RynnBrain-Trajectory
& \textbf{70.0} & 64.5 & 67.9 & 63.0 & 49.2 & 59.3 & 59.2 & 61.3 & 64.0 & 53.9 & 45.0 \\
\bottomrule
\end{tabular}
\end{adjustbox}

\label{tab:rynnbrain-9b}
\end{table}

\begin{table}[t]
\caption{Comparison with large-scale open-source and proprietary models on embodied cognition and localization benchmarks. \textnormal{* denotes results from our reproduction.}}
\centering
\setlength{\tabcolsep}{3pt}
\renewcommand{\arraystretch}{1.5}

\begin{adjustbox}{width=\textwidth}
\begin{tabular}{
p{3.5cm}
!{\color{rynn}\vrule width 1.2pt}
>{\columncolor{rynn!7}\fontsize{11.5pt}{12.6pt}\selectfont}c
!{\color{rynn}\vrule width 1.2pt}
c c c c c c c c c c
}
\toprule
\multirow{2}{*}{\textbf{Benchmark}}
& \makecell{\textbf{RynnBrain 1.1}}
& \makecell{\textbf{Cosmos 3}\\\textbf{Super}}
& \makecell{\textbf{RoboBrain}\\\textbf{2.0}}
& \makecell{\textbf{Cosmos-}\\\textbf{Reason2}}
& \textbf{Pelican-VL}
& \textbf{Qwen3.5}
& \textbf{Gemini 3 Pro}
& \textbf{GPT 5.4}
& \makecell{\textbf{Claude}\\\textbf{Sonnet 4.6}}
& \makecell{\textbf{Gemini Robotics}\\\textbf{-ER 1.5}}
& \makecell{\textbf{HY-Embodied}\\\textbf{0.5}}
\\
& \textbf{122B-A10B}
& \textbf{32B}
& \textbf{32B}
& \textbf{32B}
& \textbf{72B}
& \textbf{122B-A10B}
& \textbf{-}
& \textbf{-}
& \textbf{-}
& \textbf{-}
& \textbf{A32B}
\\
\midrule

VSI-Bench
& \textbf{75.0} & 60.9 & 42.7 & 67.0* & 57.3 & 66.6* & 48.8* & 49.2* & 44.4* & 45.8 & 68.3 \\
MMSI
& \textbf{52.0} & 41.8 & 28.5* & 32.1* & 30.7* & 9.2* & 49.2 & 37.5* & 33.0* & -    & 39.2 \\
ERQA
& 54.3 & 51.2 & 46.0 & 45.3 & 43.0 & 62.0 & \textbf{70.5} & 47.5* & 44.8* & 54.8 & 62.3 \\
RoboSpatial-Pointing
& \textbf{69.7} & 24.2* & - & 34.4 & - & 59.8 & 27.0* & 20.8* & 5.8* & 31.1 & - \\
RoboSpatial-VQA
& 70.6 & 74.1* & - & 67.1 & - & 68.9 & 66.7* & 75.0* & 55.3* & \textbf{79.3} & - \\
RoboSpatial-Overall
& 70.3 & 56.7* & \textbf{72.4} & 55.4 & 55.4 & 65.7* & 53.0* & 56.1* & 38.0* & 62.6 & - \\
MindCube
& \textbf{89.6} & 40.2* & 29.2 & 39.4* & 32.5* & 30.8* & 70.8 & 10.4* & 21.0* & 54.7 & 69.2 \\
EmbSpatial
& 83.7 & \textbf{83.8*} & 73.8* & 78.5* & 76.6 & 80.5 & 83.6 & 81.8* & 61.5* & 78.4 & - \\
RynnBrain-Object
& \textbf{77.0} & 48.3 & 26.2 & 36.8* & 42.2 & 55.4 & 58.4 & 55.5 & 33.5 & - & - \\
RynnBrain-Spatial
& \textbf{70.0} & 34.5 & 11.6 & 38.4* & 32.2 & 31.0 & 38.5 & 43.3 & 36.7 & - & - \\

RefSpatial-Bench
& \textbf{79.1} & 57.0 & 54.0 & 45.5* & 49.5 & 69.3 & 65.5 & 26.7* & 14.6* & 48.5 & 57.2 \\
RynnBrain-Grounding
& \textbf{86.6} & 74.4 & 0.0 & 60.0 & 10.8 & 66.6 & 62.1 & 24.2 & 3.3 & - & - \\
RynnBrain-Area
& 60.7 & 53.0 & 45.3 & 37.6 & 53.2 & 41.9 & \textbf{61.5} & 37.1 & 15.8 & - & - \\
RynnBrain-Affordance
& \textbf{91.3} & 84.6 & 76.1 & 83.9 & 87.3 & 79.0 & 86.0 & 83.4 & 60.0 & - & - \\
RynnBrain-Trajectory
& 69.7 & 69.3 & 60.3 & 64.0 & 64.1 & 62.8 & 72.0 & \textbf{72.8} & 56.7 & - & - \\
\bottomrule
\end{tabular}
\end{adjustbox}

\label{tab:rynnbrain-large-scale}
\end{table}

\subsection{Embodied Cognition Capability} 

We evaluate RynnBrain 1.1 on a diverse suite of embodied cognition benchmarks covering video-based spatial intelligence, multi-view reasoning, embodied question answering, robot-oriented spatial understanding, and object-centric cognition. The evaluation includes VSI-Bench~\cite{yang2025thinking}, MMSI~\cite{yang2025mmsi}, ERQA~\cite{team2025gemini}, RoboSpatial-VQA~\cite{song2025robospatial}, MindCube~\cite{yin2025spatial}, EmbSpatial~\cite{du2024embspatial}, RynnBrain-Object, and RynnBrain-Spatial. As shown in \Cref{tab:rynnbrain-2b,tab:rynnbrain-9b}, RynnBrain 1.1 improves substantially over the previous RynnBrain models on reasoning-intensive and spatially demanding tasks. At the 2B scale, the improvements are task-dependent: MMSI increases from 34.1 to 40.5, MindCube from 50.1 to 61.7, and RynnBrain-Spatial from 57.2 to 63.8, while several RoboSpatial and EmbSpatial results remain below RynnBrain-2B. The gains become more systematic at the 9B scale, where RynnBrain 1.1 outperforms RynnBrain-8B on eight of ten shared benchmarks. In particular, MindCube improves from 56.6 to 86.9, MMSI from 39.6 to 47.0, and RynnBrain-Spatial from 59.9 to 67.9, indicating stronger multi-view and spatial reasoning. Compared with other models at similar scales, RynnBrain 1.1 achieves leading results on multiple cognition benchmarks at both the 2B and 9B scales. RynnBrain 1.1-122B-A10B further extends this advantage to the large-model regime, with performance increasing from 40.5 to 47.0 and 52.0 on MMSI, and from 61.7 to 86.9 and 89.6 on MindCube, when scaling from 2B to 9B and 122B-A10B. These results show that model scaling is particularly beneficial for reasoning-intensive embodied cognition. 

\subsection{Embodied Localization Capability} We evaluate embodied localization across six spatial grounding tasks: RefSpatial-Bench~\cite{zhou2025roborefer},RoboSpatial-Pointing~\cite{song2025robospatial}, RynnBrain-Grounding, RynnBrain-Area, RynnBrain-Affordance, and RynnBrain-Trajectory. These benchmarks cover language-guided spatial reference, object and region localization, affordance grounding, and trajectory prediction. RynnBrain 1.1 improves over the previous RynnBrain models on location benchmarks at both the 2B and 9B scales. At 2B, RefSpatial-Bench increases from 52.7 to 58.5 and RynnBrain-Grounding from 79.1 to 84.1. At 9B, the gains become larger on relational and motion-oriented tasks: RefSpatial-Bench improves from 59.2 to 67.2, while RynnBrain-Trajectory increases from 64.5 to 70.0. The relatively small gain on RynnBrain-Affordance reflects its already high performance in RynnBrain 1.0. Both RynnBrain 1.1-2B and RynnBrain 1.1-9B achieve the best results among models in their respective scale groups on most of these location benchmarks. At the large scale, RynnBrain 1.1-122B-A10B achieves leading performance on RefSpatial-Bench, RynnBrain-Grounding, and RynnBrain-Affordance. The clearest scaling trend appears on RefSpatial-Bench, which improves from 58.5 at 2B to 67.2 at 9B and 79.1 at 122B-A10B, showing that relational spatial grounding benefits strongly from increased model capacity.

\subsection{Scaling Analysis}
\label{sec:scaling}

\begin{figure*}[t]
    \centering
    \includegraphics[width=\textwidth]{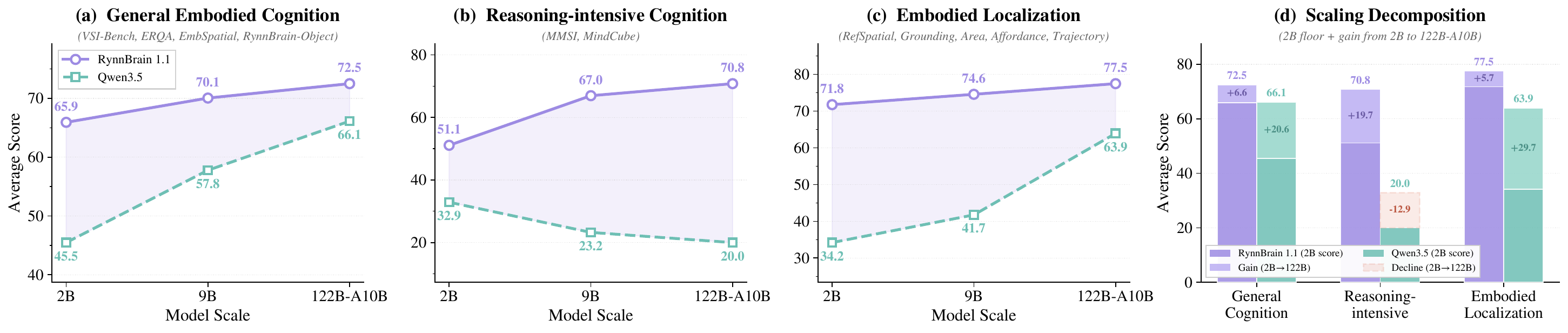}
    \caption{\textbf{Scaling analysis of RynnBrain 1.1 vs.\ Qwen3.5 from 2B to 122B-A10B.} (a)--(c) Average scores across three capability groups: general embodied cognition, reasoning-intensive cognition, and embodied localization. (d) Scaling decomposition showing the 2B performance floor and the gain from scaling to 122B-A10B.}
    \label{fig:scaling}
\end{figure*}

A central contribution of RynnBrain 1.1 is studying how embodied capability evolves with model scale under a unified training recipe. We compare RynnBrain 1.1 (2B, 9B, 122B-A10B) against the corresponding raw Qwen3.5 baselines at matched model sizes, grouping benchmarks into three categories: \textit{general embodied cognition} (visual recognition, spatial understanding, and object-centric reasoning), \textit{reasoning-intensive cognition} (multi-view spatial reasoning and video-based temporal inference), and \textit{embodied localization} (language-guided spatial grounding and trajectory prediction). Figure~\ref{fig:scaling} summarizes the scaling trends across these groups.

\paragraph{Embodied capabilities exhibit non-uniform scaling laws.}
Unlike standard vision--language benchmarks where scaling consistently improves performance, we observe three distinct scaling regimes in the embodied domain.
(1)~For \textit{general embodied cognition}, both RynnBrain 1.1 and Qwen3.5 improve monotonically with scale, and the performance gap steadily narrows. This mirrors conventional MLLM scaling---larger models reliably improve on recognition and comprehension tasks, and general model scaling can partially compensate for the lack of embodied supervision.
(2)~For \textit{reasoning-intensive cognition}, the two models diverge fundamentally: RynnBrain 1.1 improves steadily (+38.6\%), while Qwen3.5 exhibits \textit{negative scaling} ($-$39.2\%), with the gap widening from 18.2 to 50.8. This indicates that multi-view and temporal reasoning capabilities are not emergent from general VLM scaling but require explicit embodied supervision as a prerequisite. Without spatiotemporal grounding, stronger language priors in larger models may override weak visual-spatial signals, producing more confident but less accurate responses.
(3)~For \textit{embodied localization}, both models improve substantially, yet Qwen3.5 at its largest scale still underperforms RynnBrain 1.1 at its smallest. This suggests that for spatial-temporal pointing tasks, \textit{data scaling}---training on explicit coordinate supervision---is more effective than model scaling alone. RynnBrain's embodied pretraining equips even its compact model with structured spatial output capabilities that pure language modeling must recover implicitly, requiring disproportionately more parameters to approximate.

\paragraph{Embodied pretraining as a scaling enabler.}
The decomposition in Figure~\ref{fig:scaling}(d) further illustrates this contrast. RynnBrain 1.1 consistently establishes a higher performance floor across all categories, reflecting the value of embodied data even at compact scales. More critically, on reasoning-intensive tasks, embodied pretraining determines whether scaling \textit{helps or hurts}---it transforms a capability that degrades under pure VLM scaling into one that benefits consistently.
These findings indicate that for the embodied domain, data-driven spatial supervision and model scaling are complementary rather than substitutable: scaling amplifies the benefit of embodied pretraining on reasoning-heavy tasks, while embodied supervision ensures that scaling produces reliable gains rather than regression.

\subsection{3D Grounding}

We evaluate the native 3D grounding capability of RynnBrain 1.1 at both the 2B and 9B
scales. On SUN RGB-D, RynnBrain 1.1-2B achieves 34.28 AP@15, already outperforming
general-purpose VLMs such as Seed1.5-VL (33.5) and Gemini 2.0 Pro (32.5). Scaling the model
to 9B further improves the result to \textbf{41.12 AP@15}, substantially narrowing the gap
to the closed-source Gemini Robotics-ER (48.3). These results show that explicit 3D-grounded
training can equip even compact models with metric spatial understanding, while larger model
capacity further strengthens depth, size, and orientation reasoning.

A similar scaling trend is observed on WildDet3D-Bench. RynnBrain 1.1-2B obtains
17.36 AP3D, while RynnBrain 1.1-9B reaches \textbf{23.44 AP3D}, surpassing the specialized
WildDet3D detector trained with additional in-domain data (22.6). The improvement from 2B
to 9B indicates that native 3D grounding benefits from model scaling in addition to explicit
3D supervision. Overall, RynnBrain 1.1 demonstrates strong language-conditioned 3D grounding
across model sizes, with the 9B model achieving competitive performance against both
specialized detectors and substantially larger proprietary systems.
\begin{figure}[t]
    \centering
    \includegraphics[width=\linewidth]{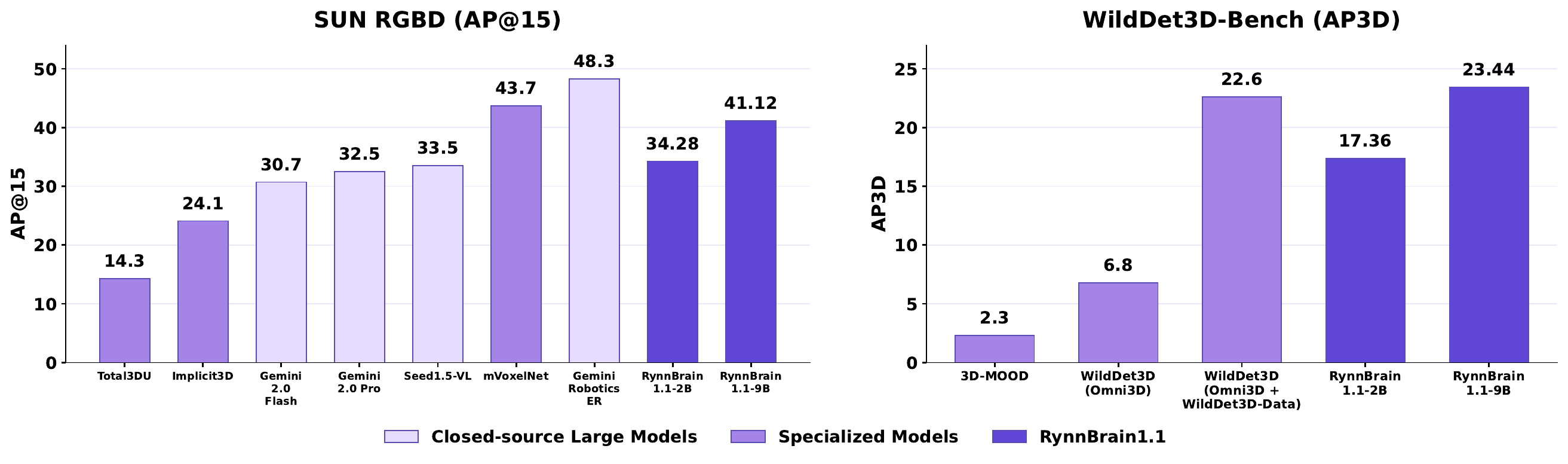}
    \caption{Performance Comparison on SUN RGB-D and WildDet3D-Bench with RynnBrain 1.1-2B and 9B.}
    \label{fig:benchmark_bar_charts}
\end{figure}

\begin{figure*}[t]
    \centering
    \includegraphics[page=1,width=\linewidth]{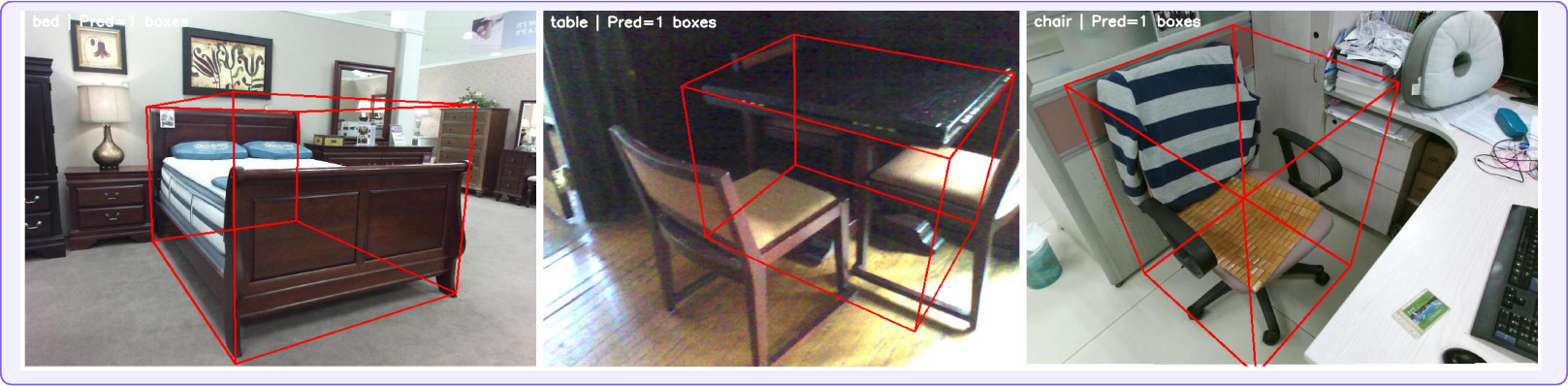}
    \caption{
    Qualitative examples of language-conditioned 3D grounding by RynnBrain 1.1.
    Given an image and a target object description, the model predicts an oriented
    3D bounding box in the camera coordinate system. The examples cover objects with
    different scales, shapes, viewpoints, and levels of occlusion in indoor environments.
    }
    \label{fig:3d-cases}
\end{figure*}

In addition to the quantitative evaluation, we visualize representative predictions in
\Cref{fig:3d-cases}. RynnBrain 1.1 can localize target objects and estimate their spatial
extent and orientation from a single image, including large furniture such as beds, tables,
and chairs. These examples illustrate the model's ability to produce physically grounded
3D predictions across diverse viewpoints and scene layouts.

\subsection{Contact Point Prediction}
\begin{figure*}[t] 
\centering 
\includegraphics[page=1,width=\linewidth]{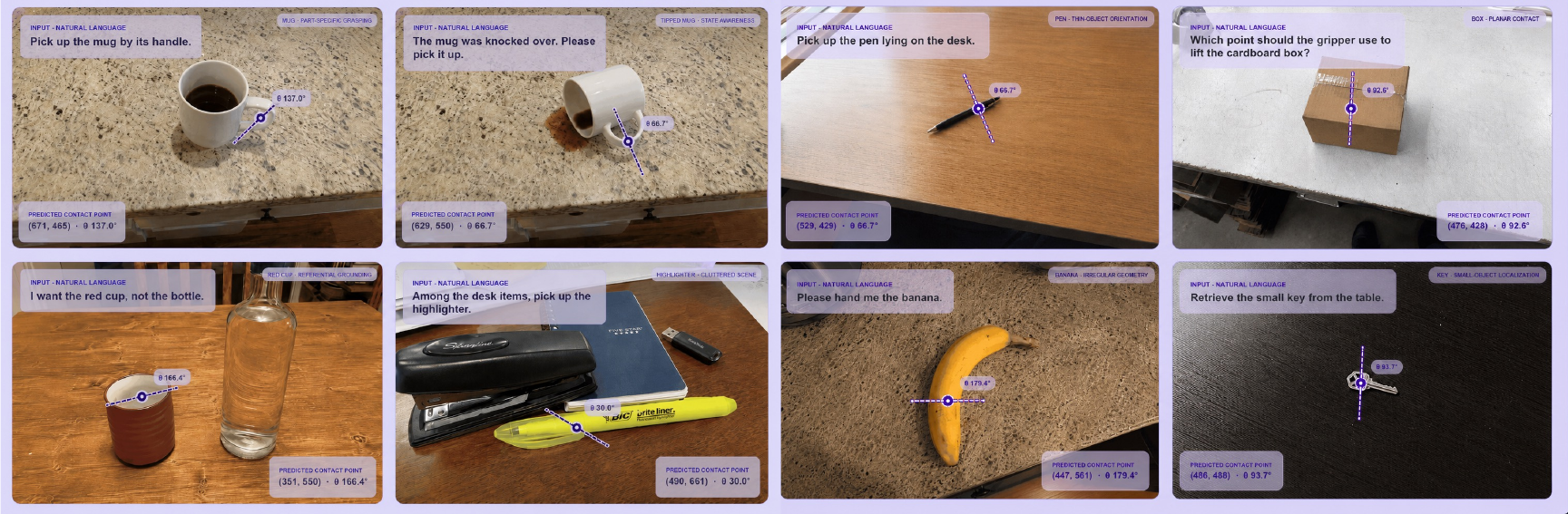} 
\caption{ Qualitative results of contact point prediction by RynnBrain 1.1. Given an image and a natural-language instruction, the model predicts an action-relevant contact point together with its in-plane grasp orientation. The examples cover diverse object geometries, object states, target parts, and cluttered scenes, demonstrating instruction-conditioned contact grounding beyond object-level localization. Coordinates are normalized to $[0,1000]$. } 
\label{fig:contact-point} 
\end{figure*}

Evaluating contact point prediction remains challenging because there is currently no standardized metric that reliably reflects the functional validity of a predicted contact. Pixel-level distance to a single annotation may penalize alternative contact points that are equally valid for manipulation, while conventional grasp-box metrics are not directly applicable to our contact-centered representation. We therefore provide a qualitative evaluation in \Cref{fig:contact-point} to illustrate the capability of RynnBrain 1.1  across diverse objects and instructions. Here we show the cases produced by RynnBrain 1.1-9B.

The visualized cases show that RynnBrain 1.1 can identify task-relevant contact locations rather than merely predicting object centers. For example, the model selects the handle when asked to pick up a mug, adapts its prediction when the mug is overturned, and produces appropriate contact locations and orientations for thin objects, boxes, and irregularly shaped items. It also distinguishes the instructed target in cluttered or referential settings. These results demonstrate that RynnBrain 1.1 can jointly reason about object identity, geometry, state, and intended interaction when generating contact-oriented spatial predictions.
\begin{figure}[t]
    \centering
    \includegraphics[page=1,width=\textwidth]{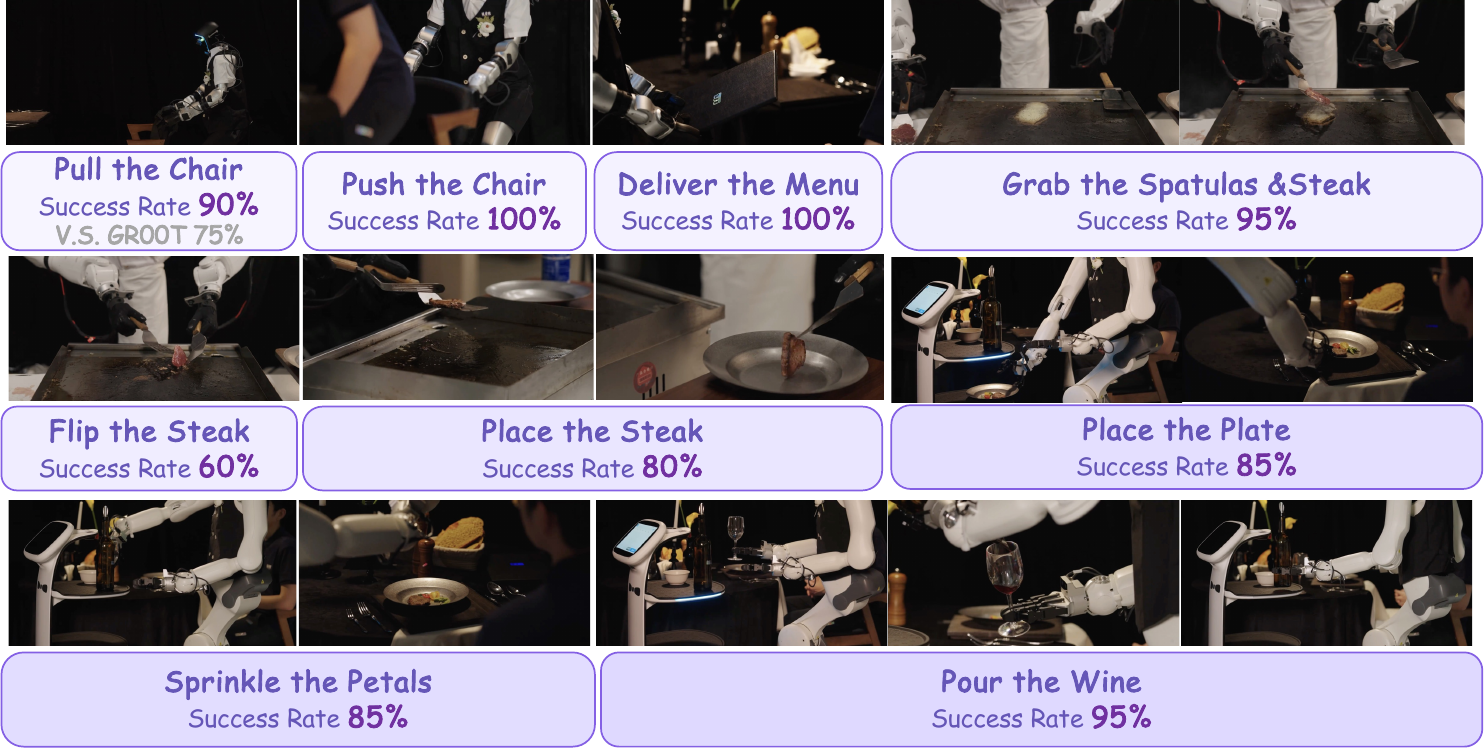}
    \caption{
    Real-robot evaluation across all tasks considered in our experiments. The tasks are
    performed in diverse real-life environments and cover whole-body manipulation, bimanual
    coordination, dexterous-hand operation, object placement, furniture interaction, and
    object delivery. Representative execution snapshots and success rates over 20 trials are
    shown for each task.
    }
    \label{fig:real-robot-tasks}
\end{figure}

\subsection{Real-Robot Evaluation}
\label{sec:realrobot}

We evaluate RynnBrain-VLA on real robots to answer three questions:
(i) can RynnBrain-VLA achieve stable performance on real-world manipulation tasks?
(ii) does a stronger embodied foundation model lead to a better VLA policy?
and (iii) can joint training on data from heterogeneous robot embodiments provide mutual benefits across embodiments?

\subsubsection{Experimental Setup}
\label{sec:realrobot_setup}

\paragraph{Evaluation Protocol.}
We deploy RynnBrain-VLA on three heterogeneous embodiments, as summarized in
\Cref{tab:embodiments_cop} and \Cref{fig:real-robot-tasks}. For each task, we conduct
20 trials with randomized initial object placements and report the final task success rate
under a fixed success criterion. Since some long-horizon tasks can be naturally decomposed
into multiple sub-tasks, final success alone does not fully reflect the execution progress of
a policy. We therefore additionally report a process score that measures the fraction of
completed sub-tasks. For a task with $K$ sub-tasks, the process score is defined as
\begin{equation}
    \mathrm{ProcessScore}
    =
    \frac{1}{20K}
    \sum_{i=1}^{20}
    \sum_{j=1}^{K}
    s_{i,j},
\end{equation}
where $s_{i,j}\in\{0,1\}$ indicates whether the $j$-th sub-task is successfully completed
in the $i$-th trial. Sprinkle the Petals, Pour the Wine, and Grab the Spatulas contain 4, 5, 3 subtasks, respectively. This metric provides a more fine-grained measure of task progress,
especially for long-horizon manipulation tasks in which a policy may complete several
intermediate stages without reaching the final goal.

\paragraph{Deployment and Action Space.}
For all tasks, inference is performed locally on a workstation equipped with a single
NVIDIA GeForce RTX 4090 GPU. The workstation is directly connected to the robot and
transmits the predicted actions for real-time execution. For tasks involving the Unitree G1
humanoid, we employ SONIC as the whole-body controller. The model predicts a 64-dimensional SONIC latent motion token alongside a 14-dimensional
dexterous-hand prediction corresponding to the G1's three-finger-joint hand, which is
represented within the unified action space's Hand group.
These two outputs are concatenated
into a 78-dimensional action representation and decoded by the SONIC controller into target
positions for the actuated joints of the Unitree G1. The 64-dimensional SONIC token is
a G1-specific output that lies outside the 81-dimensional unified action space and is used
only for the Unitree G1.
For tasks involving the Tianji-Wuji system (Tianji Arm and Wuji Hand), the policy activates
the Arm-Joint group (14D, two arms) and the Hand group (40D, two dexterous hands),
yielding a 54-dimensional active action representation that is sent directly to the
Tianji-Wuji controller. For tasks involving the Astribot S1, the policy activates the
Arm-Joint (14D), Gripper (2D), Head (3D) and Torso (4D), with the
dexterous-hand dimensions masked out.

\paragraph{Baselines and Training Settings.}
We select GR00T N1.7~\cite{gr00tn1_2025} and $\pi 0.5$~\cite{pi_0.5} as representative
generalist VLA baselines. We additionally construct Qwen-Based-VLA by applying the same VLA
post-training recipe used for RynnBrain-VLA to the Qwen base model.
For RynnBrain-VLA,
we consider two training settings. In the single-task setting, a separate policy is fine-tuned
for each individual task. In the generalist setting, we jointly fine-tune a single policy on
multi-task and multi-embodiment data using our unified action space, and denote the resulting
model as RynnBrain-VLA$_{\mathrm{Generalist}}$. To obtain the best deployment performance
for each policy, we use method-specific action-chunk lengths rather than enforcing a shared
chunk size across all models. Specifically, we use an action-chunk length of 40 for
GR00T N1.7, 50 for $\pi 0.5$, and 32 for RynnBrain-VLA, including both the single-task
and generalist settings, and Qwen-Based-VLA.

\paragraph{Tasks.} We design three challenging long-horizon manipulation tasks for quantitative comparison: \textit{Sprinkle the Petals}, \textit{Pour the Wine}, and \textit{Grab the Spatulas}. Each task consists of multiple temporally ordered sub-tasks and requires the policy to complete a sequence of perception, reaching, grasping, manipulation, and placement operations. The task setups are illustrated in \Cref{fig:real-robot-tasks}. The first two tasks are performed on the Astribot robot with gripper-based whole-body control, where the robot needs to actively change its viewpoint during execution. In \textit{Sprinkle the Petals}, the policy must continuously observe the container and determine whether the petals have been fully poured out before proceeding, requiring closed-loop visual observation and task-progress understanding. \textit{Pour the Wine} requires the robot to precisely align the bottle opening with the target glass and execute a controlled pouring motion, placing stronger demands on fine-grained affordance understanding and precise manipulation. \textit{Grab the Spatulas} is performed with Tianji-Wuji dexterous hands and requires fine-grained grasping and hand control to reliably manipulate the tools. Together, these tasks cover complementary challenges in long-horizon reasoning, visual feedback, precise affordance-guided manipulation, and dexterous interaction, and serve as the main benchmark for comparing different VLA methods and training settings. Beyond the three main tasks, we further evaluate RynnBrain-VLA on a series of additional robot tasks across heterogeneous embodiments, including object placement, flipping, furniture interaction, and object delivery. These evaluations broaden the coverage of embodiments and manipulation primitives and provide an additional assessment of policy generality.

\begin{table}[t]
\caption{Real-robot evaluation results. Each task reports the process score and final success
rate (\%). The process score measures the average fraction of completed sub-tasks, while a
trial is counted as successful only when all sub-tasks of the corresponding task are completed.}
\label{tab:embodiments_cop}
\centering
\setlength{\tabcolsep}{5pt}
\renewcommand{\arraystretch}{1.25}

\begin{tabular}{c cc cc cc cc}
\toprule
& \multicolumn{2}{c}{\textbf{Sprinkle the Petals}}
& \multicolumn{2}{c}{\textbf{Pour the Wine}}
& \multicolumn{2}{c}{\textbf{Grab the Spatulas}}
& \multicolumn{2}{c}{\textbf{Average}} \\
\textbf{Model}

& \multicolumn{2}{c}{\textit{Astribot }}
& \multicolumn{2}{c}{\textit{Astribot }}
& \multicolumn{2}{c}{\textit{Tianji-Wuji}}
& \multicolumn{2}{c}{--} \\

\cmidrule(lr){2-3}
\cmidrule(lr){4-5}
\cmidrule(lr){6-7}
\cmidrule(lr){8-9}

& \textbf{Proc.} & \textbf{Succ.}
& \textbf{Proc.} & \textbf{Succ.}
& \textbf{Proc.} & \textbf{Succ.}
& \textbf{Proc.} & \textbf{Succ.} \\
\midrule

Qwen-Based-VLA
& 75.00 & 65.00
& 75.00 & 65.00
& 55.00 & 50.00
& 68.33 & 60.00 \\

GR00T N1.7
& 86.25 & 75.00
& 77.00 & 65.00
& 86.67 & 80.00
& 83.31 & 73.33 \\

$\pi_{0.5}$
& 65.00 & 60.00
& 69.00 & 65.00
& 83.33 & 70.00
& 72.44 & 65.00 \\

\midrule

\rowcolor{rynn!7}
RynnBrain-VLA
& 87.50 & \textbf{85.00}
& 88.00 & 80.00
& \textbf{98.33} & \textbf{95.00}
& 91.28 & 86.67 \\

\rowcolor{rynn!7}
RynnBrain-VLA$_{\mathrm{Generalist}}$
& \textbf{88.75} & \textbf{85.00}
& \textbf{97.00} & \textbf{95.00}
& 96.67 & \textbf{95.00}
& \textbf{94.14} & \textbf{91.67} \\

\bottomrule
\end{tabular}
\end{table}

\subsubsection{Stable Real-World Performance}
As shown in \Cref{fig:real-robot-tasks} and \Cref{tab:embodiments_cop}, RynnBrain-VLA achieves stable performance across multiple robot embodiments, including the Astribot bimanual platform, the Tianji-Wuji dexterous-hand system, and the Unitree G1 humanoid. The policy performs reliably under substantially different action spaces and control interfaces, covering gripper-based whole-body manipulation, dexterous grasping, bimanual coordination, and humanoid whole-body control. In particular, the Astribot tasks require the robot to actively change its viewpoint during execution and continuously update its actions from visual feedback. The strong results on these tasks indicate that RynnBrain-VLA remains robust to viewpoint changes rather than relying on a fixed observation configuration. 

RynnBrain-VLA also performs well on the Unitree G1 humanoid. On \textit{Pull the Chair}, which involves coordinated humanoid whole-body motion, RynnBrain-VLA achieves a 90\% success rate, compared with 75\% for GR00T N1.7 combined with the same SONIC controller. Since GR00T N1.7 with SONIC is a natural baseline for predicting SONIC motion tokens, this comparison provides additional evidence that RynnBrain-VLA can produce reliable actions even for challenging whole-body control tasks. Overall, the results demonstrate that its real-world performance is stable not only across tasks, but also across robot morphologies, viewpoints, and control configurations.
\subsubsection{A Stronger Embodied Base Yields a Better VLA} We next investigate whether an embodied foundation model provides a stronger initialization for VLA post-training. RynnBrain-VLA and Qwen-Based-VLA follow the same VLA post-training recipe and use the same action-chunk length, while being initialized from RynnBrain and Qwen Based Model, respectively. 
As shown in \Cref{tab:embodiments_cop}, RynnBrain-VLA consistently outperforms Qwen-Based-VLA across all three long-horizon tasks. Its average process score improves from 68.33\% to 91.28\%, while the average final success rate increases from 60.00\% to 86.67\%. The advantage is particularly pronounced on \textit{Grab the Spatulas}, where RynnBrain-VLA achieves a 95\% success rate compared with 50\% for Qwen-Based-VLA. These results suggest that the capabilities acquired during embodied foundation-model training transfer effectively to downstream VLA post-training. Compared with directly post-training a general-purpose VLM as a VLA policy, initializing from RynnBrain leads to more reliable execution of multi-stage manipulation tasks and substantially fewer failures during intermediate stages. This comparison provides direct evidence that a stronger embodied base model can serve as a better foundation for real-world VLA policies.

\subsubsection{Cross-Embodiment Generalization} \label{sec:realrobot_cross} To examine whether the capability of RynnBrain-VLA is tied to a specific robot embodiment, we adapt it to three morphologically and mechanically distinct robot platforms through the deployment framework described in \Cref{sec:vla_deploy}. As shown in \Cref{tab:embodiments_cop} and \Cref{fig:real-robot-tasks}, RynnBrain-VLA achieves stable performance across humanoid robots, dexterous-hand manipulation, whole-body control, and bimanual grasping settings. These embodiments differ substantially in morphology, action dimensionality, and low-level control interfaces. The consistent performance across these settings suggests that the embodied representations learned by RynnBrain capture task- and interaction-level knowledge that is not tightly coupled to a particular robot morphology or action parameterization, allowing the same embodied foundation model to serve as a shared starting point for heterogeneous robot policies. 

More interestingly, we find that jointly training multiple tasks and embodiments within a single VLA does not introduce obvious interference. With the unified action-space  strategy in \Cref{sec:unified_action_space}, RynnBrain-VLA$_{\mathrm{Generalist}}$ improves the average process score from 91.28\% to 94.14\% and the final success rate from 86.67\% to 91.67\% compared with separately fine-tuned single-task policies. We hypothesize that different embodiments share common interaction structures---such as object affordances, task progress, and reach--grasp--manipulate patterns---even though their executable actions are different. Joint training therefore allows these shared structures to reinforce a common visual--action representation, while unified action-space  preserves embodiment-specific control. This suggests that cross-embodiment data can act as complementary supervision: the key benefit of a unified VLA may lie not only in supporting more robots with one model, but also in allowing different embodiments to teach each other how the physical world can be interacted with.

\section{Conclusion and Future Work}

We presented \textbf{RynnBrain 1.1}, an embodied foundation model family spanning
2B, 9B, and 122B-A10B scales. RynnBrain 1.1 studies embodied capability scaling under
a unified training framework and demonstrates clear improvements in multi-view reasoning,
spatial grounding, and robot-oriented understanding. We further introduce explicit 3D-grounded
supervision for the 2B and 9B models, enabling native language-conditioned 3D grounding,
together with a compact contact-centered representation for interaction grounding.

We also develop RynnBrain-VLA to connect embodied understanding with real-world action. Under the same post-training recipe, RynnBrain-VLA substantially outperforms a Qwen-based VLA on long-horizon manipulation tasks. Through a unified action space with embodiment-specific masking, the policy is deployed across a Unitree G1 humanoid, an Astribot bimanual robot, and a Tianji-Wuji dexterous-hand system. Joint multi-task and multi-embodiment training further improves real-robot performance, providing initial evidence that heterogeneous robot data can contribute complementary supervision to a shared policy.

Looking forward, we aim to extend RynnBrain toward a unified multimodal model that combines perception, understanding, generation, spatial grounding, and action within a common framework. An important direction is to develop RynnBrain into the cognitive core of a general embodied agent, integrating long-term memory, world modeling, task planning, active perception, and closed-loop interaction. Such an agent could continuously reason about changes in the physical world, generate and revise plans from feedback, and acquire new skills through interaction across diverse tasks and robot embodiments.

\bibliographystyle{assets/plainnat}
\bibliography{paper}

\end{document}